\documentclass[runningheads]{llncs}

\usepackage{eccv}

\usepackage{eccvabbrv}

\usepackage{url}% === added by kunming
\usepackage{enumitem}
\usepackage{comment}
\usepackage{times}
\usepackage{multirow}
\usepackage{multicol}
\usepackage{booktabs}
\usepackage{bbding} % add Checkmark
\usepackage{float}
\usepackage{subcaption}
\usepackage{tabularx}
\usepackage{array}
\usepackage{mathtools}% === added by kunming
\usepackage{color, colortbl}
\usepackage[dvipsnames]{xcolor}
\usepackage{wrapfig}% added by suyi

\definecolor{highlight}{RGB}{236, 236, 236}

\newcommand{\ourname}{\color{myBlack}{PointRegGPT}}
\newcommand{\ourdiffusionname}{\color{myBlack}{diffusion model}}
\newcommand{\ourdepthcorretionname}{\color{myBlack}{depth correction module}}
\newcommand{\ourdepthaugmentationname}{\color{myBlack}{depth augmentation module}}
\usepackage{graphicx}
\usepackage{booktabs}

\usepackage[accsupp]{axessibility}  % Improves PDF readability for those with disabilities.

\usepackage{hyperref}

\usepackage{orcidlink}

\definecolor{myPurple}{rgb}{0.4, .0, .8}
\definecolor{myGreen}{rgb}{0, .8, .3}
\definecolor{myRed}{rgb}{0.8, .2, .2}
\definecolor{myPink}{rgb}{0.8, .5, .5}
\definecolor{myOrange}{rgb}{0.7, 0.45, 0.2}
\definecolor{myBlue}{rgb}{.0, .0, 1.0}
\definecolor{myBlue2}{rgb}{.0, .0, 0.5}
\definecolor{myBlack}{rgb}{.0, .0, 0.0}
\definecolor{mycyan}{rgb}{.39,.58,.93}

\begin{document}

\title{\ourname: Boosting 3D Point Cloud Registration \\ using Generative Point-Cloud Pairs for Training}

\titlerunning{\ourname}

\author{Suyi Chen\inst{1,4}\orcidlink{0009-0009-2566-7526} \and
Hao Xu\inst{2}\orcidlink{0000-0003-3676-5737} \and
Haipeng Li\inst{1}\orcidlink{0000-0003-3983-9287} \and
Kunming Luo\inst{3}\orcidlink{0000-0002-5070-7392} \and
Guanghui Liu\inst{1}\orcidlink{0000-0002-4170-4552} \and
Chi-Wing Fu\inst{2}\orcidlink{0000-0002-5238-593X} \and
Ping Tan\inst{3}\orcidlink{0000-0002-4506-6973} \and
Shuaicheng Liu\inst{1,4}\thanks{Corresponding author.}\orcidlink{0000-0002-8815-5335}
}

\authorrunning{S.~Chen, H.~Xu, et al.}

\institute{University of Electronic Science and Technology of China \and
The Chinese University of Hong Kong \and
Hong Kong University of Science and Technology \and
Megvii Technology}

\maketitle

\footnotetext[2]{Department of Computer Science and Engineering; Institute of Medical Intelligence and XR.}
\footnotetext[3]{HKUST Shenzhen-Hong Kong Collaborative Innovation Research Institute.}

\begin{abstract}

Data plays a crucial role in training learning-based methods for 3D point cloud registration.
However, the real-world dataset is expensive to build, while rendering-based synthetic data suffers from domain gaps.
In this work, we present \textbf{\ourname{}}, boosting 3D \textbf{Point} cloud \textbf{Reg}istration using \textbf{G}enerative \textbf{P}oint-cloud pairs for \textbf{T}raining.
Given a single depth map, we first apply a random camera motion to re-project it into a target depth map. 
Converting them to point clouds gives a training pair.
To enhance the data realism, we formulate a generative model as a depth inpainting diffusion to process the target depth map with the re-projected source depth map as the condition.
Also, we design a depth correction module to alleviate artifacts caused by point penetration during the re-projection.
To our knowledge, this is the first generative approach that explores realistic data generation for indoor 3D point cloud registration.
When equipped with our approach, several recent algorithms can improve their performance significantly and achieve SOTA consistently on two common benchmarks.
The code and dataset will be released on {\small \url{https://github.com/Chen-Suyi/PointRegGPT}}.


\end{abstract}    
\section{Introduction}
\label{sec:intro}
\begin{figure}[h]
    \centering
    \includegraphics[width=0.75\linewidth]{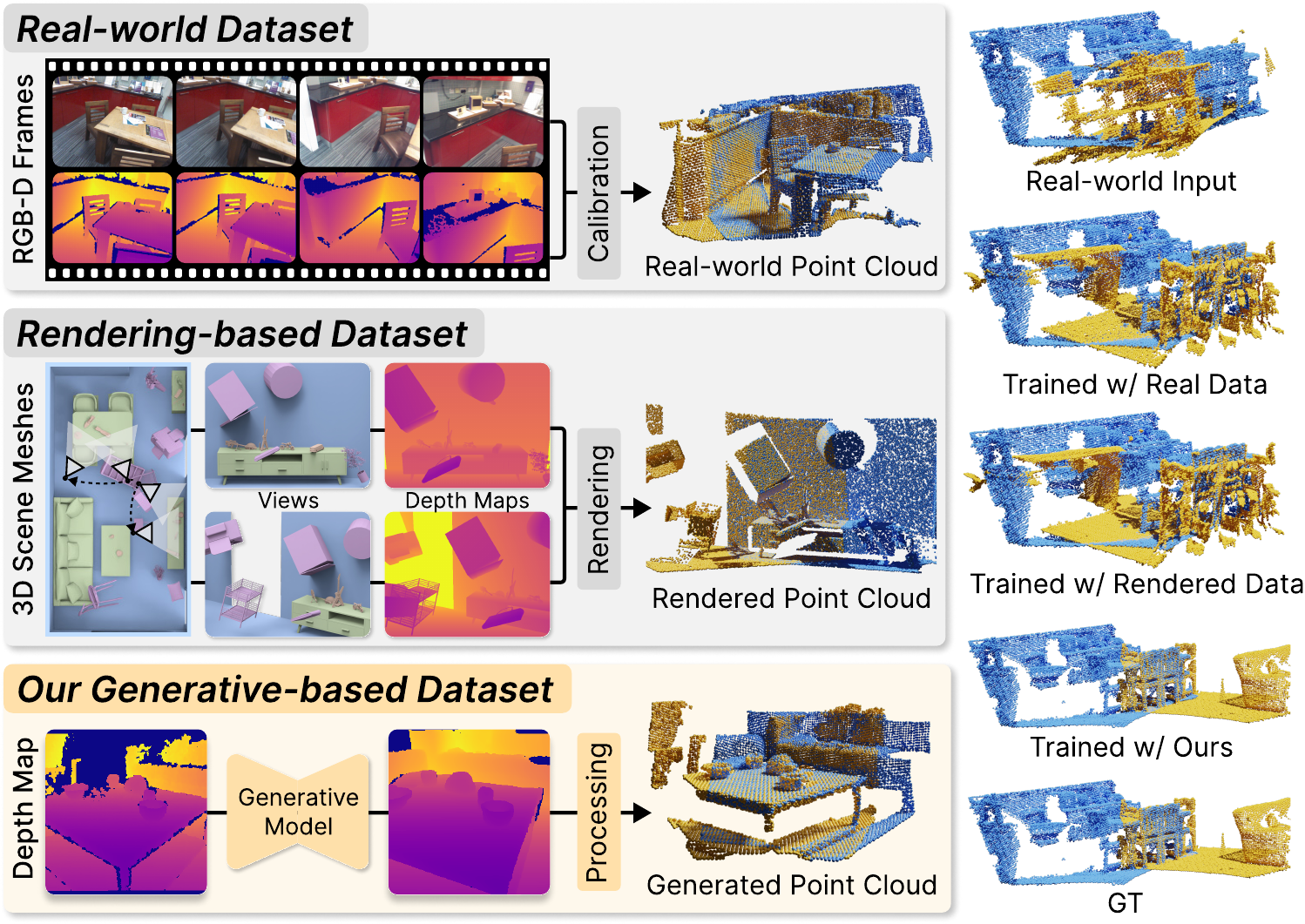}
    \caption{Comparison among methods of obtaining training data for 3D point cloud registration. Our approach stands out by generating overlapping point clouds from single depth maps, bypassing the need for laborious human annotations in real-world datasets or additional 3D scene meshes required in rendering-based datasets.}
    \label{fig: teaser}
\end{figure}

3D point cloud registration is a fundamental task with various applications in robotics. 
Recent learning-based registration methods~\cite{choy2019fully, bai2020d3feat, huang2020predator, yu2021cofinet, qin2022geometric, wang2023roreg} have achieved significant success, largely driven by the utilization of large-scale datasets. 
The cornerstone of these approaches lies in the quality of their training datasets.

To build robust models for real-world scenarios, a qualified dataset should possess three key attributes: data quantity, label quality, and point cloud realism.
However, it is challenging for existing point cloud registration datasets to simultaneously satisfy all these characteristics. 
The most common approach involves collecting data from real-world scenes using depth cameras and gyroscopes~\cite{zeng20173dmatch}. 
Nevertheless, gyroscope data is often prone to errors, necessitating manual annotations and complex calibration processes to ensure accurate camera pose labels. 
This limitation restricts the availability of high-quality annotated data in large quantities.
An alternative method~\cite{chen2023sira} entails rendering registration data from graphic models, enabling automatic generation of large-scale point cloud pairs with accurate camera pose labels.
However, such rendering-based synthetic dataset falls short in satisfying the realism criteria, affecting the performance of models due to the domain gap between synthetic and real-world scenes. 
Though recent efforts~\cite{chen2023sira} have attempted domain adaptation through GAN-based techniques, bridging this distribution gap remains challenging due to the inherent instability of unpaired GAN-based learning processes.

To the best of our knowledge, there is currently no dedicated method proposed for the automated generation of large-scale, high-quality, real-world point cloud registration data. 
While some efforts~\cite{wang2019prnet, yew2020-RPMNet, horache20213d} have attempted to generate point cloud registration data from individual real-world point clouds to facilitate cross-domain transfer or self-supervised learning, these methods primarily involve a relatively simple pipeline of cropping portions from individual real-world point clouds and randomly introducing camera poses to create registration data pairs. 
While the source point cloud originates from real-world acquisitions, the resulting target point clouds lack realism, leading to limited dataset diversity and generalization capability of point cloud registration.

In this paper, we introduce a novel method, \ourname{}, to automatically create large-scale, realistic datasets for enhancing 3D point cloud registration in real-world applications. \cref{fig: teaser,tab: comparison_datasets} show an illustration of our \ourname{} in comparison with existing methods of obtaining real-world dataset and rendering-based dataset.
Our approach generates point cloud pairs from single depth maps, using generative models to attain a partially overlapped target depth map under random camera movements.
This process, however, faces challenges in both (i) maintaining both 3D geometric consistency and realism during generation and (ii) avoiding artifacts caused by the ``point penetration problem'', where changes in viewpoint can lead to empty and inaccurate pixels in the target map.
To overcome challenge (i), we design a diffusion-based pipeline that consists of the re-projection module and the depth generation module, allowing for the generation of realistic depth maps from new perspectives while keeping 3D geometric consistency.
Furthermore, for challenge (ii), we learn a depth correction module that identifies and corrects regions affected by penetrated points, ensuring the generation of accurate depth maps and thereby facilitating the creation of valid point-cloud pairs.
\begin{table}[t]
    \centering
    \caption{Comparison among different indoor scene datasets. As a real-world dataset, 3DMatch~\cite{zeng20173dmatch} requires heavy human labor for annotation and calibration, while the ground-truth labels from gyroscopes are error-prone, resulting in a limited quantity of data. As a rendering-based dataset, FlyingShapes~\cite{chen2023sira} render a huge quantity of point clouds from 3D assets, which yet need human labor from professional designers. Our method can generate large-scale datasets easily with high quality without human labor.}
    \resizebox{0.9\linewidth}{!}{
        \begin{tabular}{l|c|c|c|c|c|c|c}
        \toprule
        \multirow{2}{*}{Datasets} & \multirow{2}{*}{Scene Type} & \multirow{2}{*}{\#Fragments} & \multirow{2}{*}{\#Pairs} & \multirow{2}{*}{Label Accuracy} & w/o Human & w/o Professional & \multirow{2}{*}{Realism}\\
        & & & & & Annotation & Designer & \\
        \midrule 
        3DMatch~\cite{zeng20173dmatch} & real-captured & 3,254 & 20,642 & error-prone & \XSolidBrush & \Checkmark & \Checkmark\\
        FlyingShapes~\cite{chen2023sira} & rendering & 21,550 & 107,641 & accurate & \Checkmark & \XSolidBrush & \XSolidBrush \\
        \midrule 
        Ours & generative & 320,000 & 160,000 & accurate & \Checkmark & \Checkmark & \Checkmark \\
        \bottomrule
    \end{tabular}}
    \label{tab: comparison_datasets}
\end{table}

By applying \ourname, we can generate large amount of training data from depth maps captured in the real world. 
We conduct extensive experiments on multiple point cloud registration benchmarks, demonstrating significant improvements over previous methods. 
Our contributions can be summarized as follows:
\begin{itemize}
\item We propose a new framework that can automatically generate realistic training data for 3D point cloud registration from depth maps captured in real-world scenes.
\item We design a diffusion-based pipeline, including the depth generation and depth correction modules, to 
effectively produce generative point-cloud pairs for training.
\item Extensive qualitative and quantitative experiments on multiple registration benchmarks showcase the state-of-the-art performance of our method. Notably, we leverage our generated dataset to train different registration networks and observe consistent notable improvements.
\end{itemize}
\section{Related Work}
\paragraph{\bf 3D Point Cloud Registration}
focuses on aligning point cloud pairs by estimating 3D rigid transforms, facing challenges due to partiality. Methods are categorized into direct registration and correspondence-based approaches. Direct registration includes regressing transformations from point cloud features, shown in works like ~\cite{aoki2019pointnetlk, huang2020feature, xu2021omnet, xu2022finet}, and using differentiable weighted SVD for transformation derivation, as in ~\cite{wang2019deep, wang2019prnet, idam, yew2020-RPMNet, fu2021robust}, which yet struggle with scalability and generalization in complex scenes ~\cite{huang2020predator}. 

Correspondence-based methods evolved from traditional techniques using handcrafted features~\cite{besl1992method, rusinkiewicz-normal-sampling, segal2009generalized, yang2013go, rusinkiewicz2019symmetric, FPFH} and RANSAC~\cite{fischler1981random} to deep learning approaches for robust feature descriptors~\cite{choy2019fully, gojcic2019perfect, bai2020d3feat, ao2021spinnet, poiesi2022learning, wang2022you, yu2021cofinet, yew2022regtr, qin2022geometric, wang2023roreg, yangone, lepard2021, Ao_2023_CVPR, yu2023peal,  Mei_2023_CVPR, Liu_2023_ICCV, Chen_2023_ICCV, liu2023regformer}, including deep robust estimators and carefully-designed strategies for transformation accuracy~\cite{bai2021pointdsc, choy2020deep, pais20203dregnet, Jiang_2023_CVPR, Zhang_2023_CVPR, Hatem_2023_ICCV}.

The scarcity of training data limits progress in this field. Dang \etal~\cite{Dang_2023_ICCV} and Chen \etal~\cite{chen2023sira} attempt to generate synthetic data, but face limitations like applicability to only object-level tasks or domain gaps. This prompts our use of generative models to create diverse, realistic, large-scale datasets from real data. Our experimental approach builds on high-performance methods such as PREDATOR \cite{huang2020predator}, CoFiNet \cite{yu2021cofinet}, and GeoTransformer (GeoTrans) \cite{qin2022geometric}.

% \vspace{-4mm}
\paragraph{\bf Diffusion Models}
consist of a forward process adding noise to an image and a reverse process of image recovery using neural networks, grounded in thermodynamic principles~\cite{sohl2015deep}. These models, especially DDPM~\cite{ho2020denoising}, employ score-based methods and stochastic differential equations for data generation. Enhancements like DDIM~\cite{song2020denoising} have improved efficiency. Advances include classifier-guided~\cite{dhariwal2021diffusion, liu2023more} and classifier-free~\cite{ho2022classifier} methods for controlled data generation, with techniques like LDM~\cite{rombach2022high} using cross-attention mechanisms, and DDNM~\cite{wang2022zero} utilizing range-null space decomposition. Applications of diffusion models cover image super-resolution~\cite{saharia2022image}, restoration~\cite{luo2023image, zhou2024recdiffusion, yang2024single, li2024dmhomo}, human motion prediction~\cite{tevet2022human}, and medical imaging~\cite{song2021solving}, extending to 3D generation~\cite{liu2023zero} and reconstruction~\cite{lei2023rgbd2, xu2024handbooster}. This paper pioneers their applications in 3D point cloud registration, a novel and promising area.
\section{Method}
\begin{figure*}[t]
    \centering
    \includegraphics[width=\linewidth]{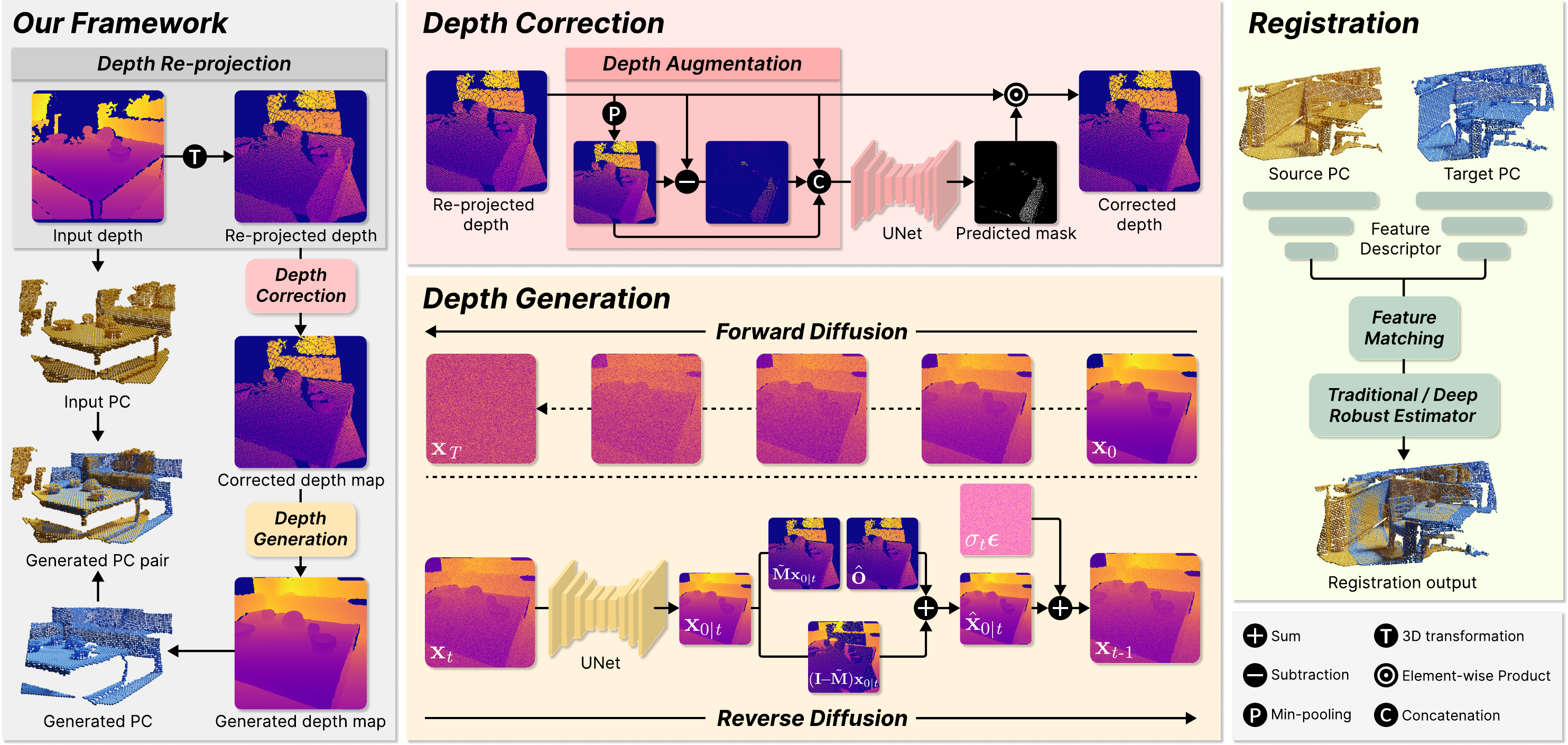}
    \caption{Our \ourname{} framework: (i) the given depth map is processed by re-projection, depth correction, and depth generation to produce a new depth map, which is then converted into a pair of partially-overlapped point clouds; (ii) depth generation utilizes a pre-trained diffusion model to generate new content while it keeps geometric consistency in the overlap region; (iii) depth correction is used to correct the unnatural depth values generated by penetrated points during re-projection, where the depth augmentation module enhances the ability to detect the wrong values; (iv) the generated dataset is used to train registration models.}
    \label{fig: pipeline}
\end{figure*}
\subsection{Overview}
Recent learning-based methods have achieved impressive performance in 3D point cloud registration, however being hindered by the scarcity of existing training data. Some previous works~\cite{Dang_2023_ICCV, chen2023sira} resort to additional synthetic data but face different limitations. To this end, we resort to diffusion models, which can learn from existing real data directly and generate diverse realistic depth maps, to build our data generation framework.
\cref{fig: pipeline} shows the framework of our method. Given a real depth map, we re-project it with a randomly-sampled pose and then inpaint it using a pre-trained diffusion model (Sec.~\ref{sec: 3.2}). Then, to avoid the artifacts caused by the point penetration problem, we design a depth correction module to eliminate the unnatural depth values from penetrated points in the re-projected depth map (Sec.~\ref{sec: 3.3}).
Finally, the original and the generated depth maps will be converted to the source point cloud and the target point cloud, respectively, which can be used as a piece of training data for point cloud registration. The generated data will be compiled into a large-scale realistic dataset to boost learning-based methods (Sec.~\ref{sec: 3.4}).

\subsection{Depth Generation}
\label{sec: 3.2}
To generate reliable training data for 3D point cloud registration, one of the challenges is to strictly guarantee the 3D geometric consistency of the overlap region between two partially overlapping point clouds.
Otherwise, the ground-truth labels could be incorrect. In some previous works that generate 3D assets through novel view synthesis~\cite{liu2023zero}, images and corresponding poses are used as conditions for guidance, which are usually embedded while processing cross-attention. However, the novel view generated in this way can hardly finely accord with the 3D geometric consistency we need. There are still inevitable failure cases due to the unpredictable diffusion process. 
To address this problem, we utilize the 3D structures behind depth maps by re-projection and inpainting. In this way, the 3D geometric consistency in overlap regions can be absolutely guaranteed by the remanent portion of the given depth map.

\paragraph{\bf Re-projection for Consistency.} 
Given a depth map $\mathbf{I} \in \mathbb{R}^{H \times W}$ and its corresponding intrinsic matrix $\mathbf{K} \in \mathbb{R}^{3 \times 3}$, we first convert it into a point cloud $\mathbf{P} \in \mathbb{R}^{N\times3}$ and transform 
it with a random sampled pose $\mathbf{T} = \{\mathbf{R}, \mathbf{t}\} \in SE(3)$: 
\begin{equation}
    \mathbf{P}^{'} = \mathbf{P}\mathbf{R}^T + \mathbf{t}
\end{equation}
where $\mathbf{R} \in SO(3)$ and $\mathbf{t} \in \mathbb{R}^3$ denote the rotation and the translation, respectively. Then, $\mathbf{P}^{'}$ is re-projected back to the pixel coordinate. The transformed point cloud will not exactly occupy every pixel again due to the following reasons: (i) part of the transformed point cloud $\mathbf{P}^{'}$ is out of the field of view, (ii) several points in the same ray happen to fall into the one pixel, and (iii) the sparsity of the point cloud leading to empty pixels. Thus, re-projection usually yields an incomplete depth map $\hat{\mathbf{O}} \in \mathbb{R}^{H \times W}$.
The incomplete depth map $\hat{\mathbf{O}}$ is then used as the condition for a pre-trained diffusion model, which can generate new contents by inpainting empty pixels of the incomplete depth map, yielding a complete depth map $\mathbf{O} \in \mathbb{R}^{H \times W}$.

\paragraph{\bf Generation for New Contents.} 
It is known that diffusion models iteratively add random noises at each time step $t$ to gradually turn a clean image $\mathbf{x}_0 \sim q \left( \mathbf{x} \right)$ to total random noise $\mathbf{x}_T \sim \mathcal{N} \left( \mathbf{0}, \mathbf{I} \right)$ during the $T$-step forward process, 
\begin{equation}
    \mathbf{x}_{t-1} = \sqrt{1 - \beta_t} \mathbf{x}_{t-1} + \sqrt{\beta_t} \boldsymbol{\epsilon}, \quad \boldsymbol{\epsilon} \sim \mathcal{N}\left( \mathbf{0}, \mathbf{I} \right),
\end{equation}
where $\beta_t$ is the predefined variance for noise and $\mathcal{N}$ denotes the Gaussian distribution; diffusion models iteratively sample $\mathbf{x}_{t-1}$ at each time step $t$ from $p \left( \mathbf{x}_{t-1} | \mathbf{x}_t, \mathbf{x}_0 \right)$ to generate image $\mathbf{x}_0 \sim q \left( \mathbf{x}_0 \right)$ from random noise $\mathbf{x}_T \sim \mathcal{N} \left( \mathbf{0}, \mathbf{I} \right)$ during the $T$-step reverse process. To finely control the output result $\mathbf{x}_0$ of a diffusion model, inspired by DDNM~\cite{wang2022zero}, we formulate the relation between the input incomplete re-projected depth map $\hat{\mathbf{O}}$ and the output complete target depth map $\mathbf{O}$ with new content as a task of image restoration. In other words, the incomplete depth map $\hat{\mathbf{O}}$ is considered as a degradation from the complete depth map $\mathbf{O}$ by eliminating some of the values using a mask $\mathbf{M} \in \{0, 1\}^{H \times W}$, \ie,
\begin{equation}
    \hat{\mathbf{O}} = \mathbf{M} \odot \mathbf{O},
\end{equation}
where $\odot$ denotes the element-wise product. For a better explanation, we re-write it in another form as below:
\begin{equation}
\label{eq: 2}
    \hat{\mathbf{o}} = \tilde{\mathbf{M}} \mathbf{o},
\end{equation}
where $\hat{\mathbf{o}} \in \mathbb{R}^{HW}$ and $\mathbf{o} \in \mathbb{R}^{HW}$ are the vectorized form of $\hat{\mathbf{O}}$ and $\mathbf{O}$, respectively; and $\tilde{\mathbf{M}} = \mathrm{diag}(\mathrm{vec}(\mathbf{M}))$ is a diagonal matrix.
Now we resort to diffusion models to predict $\mathbf{o}$. A prediction $\mathbf{x}_0$ of the diffusion model will be satisfying if it conforms well to the following two constraints: 
\begin{equation}
    Consistency: \tilde{\mathbf{M}} \mathbf{x}_0 = \hat{\mathbf{o}}, 
    \label{eq: consistency}
\end{equation}
\begin{equation}
    \text{and} \, Realism: \mathbf{x}_0 \sim q \left( \mathbf{o} \right). \label{eq: realness}
\end{equation}
\cref{eq: realness} is spontaneously satisfied by the diffusion model when it is well-trained to learn $q \left( \mathbf{o} \right)$. Thus, what we need to do is to guarantee \cref{eq: consistency} during the reverse diffusion process.
By applying range-null space decomposition, $\mathbf{x}_0$ is decomposed to the range space and the null space of $\tilde{\mathbf{M}}$:
\begin{equation}
    \mathbf{x}_0 = \tilde{\mathbf{M}}^{\dagger} \tilde{\mathbf{M}} \mathbf{x}_0 + \left( \mathbf{I} - \tilde{\mathbf{M}}^{\dagger} \tilde{\mathbf{M}} \right) \mathbf{x}_0,
\end{equation}
where $ \tilde{\mathbf{M}}^{\dagger} \tilde{\mathbf{M}} \mathbf{x}_0 $ and $\left( \mathbf{I} - \tilde{\mathbf{M}}^{\dagger} \tilde{\mathbf{M}} \right) \mathbf{x}_0$ are the projection of $\mathbf{x}_0$ in the range space and the null space of $\tilde{\mathbf{M}}$, respectively.
Theoretically, it can always satisfy \cref{eq: consistency} if the range space $\tilde{\mathbf{M}}^{\dagger} \tilde{\mathbf{M}} \mathbf{x}_0$ is replaced by $\tilde{\mathbf{M}}^{\dagger} \hat{\mathbf{o}}$, due to the fact that
\begin{equation}
    \begin{aligned}
        \tilde{\mathbf{M}} \hat{\mathbf{x}}_0 & = \tilde{\mathbf{M}} \tilde{\mathbf{M}}^{\dagger} \hat{\mathbf{o}} + \tilde{\mathbf{M}} \left( \mathbf{I} - \tilde{\mathbf{M}}^{\dagger} \tilde{\mathbf{M}} \right) \mathbf{x}_0 \\
        & = \tilde{\mathbf{M}} \tilde{\mathbf{M}}^{\dagger} \tilde{\mathbf{M}} \mathbf{o} + \tilde{\mathbf{M}} \left( \mathbf{I} - \tilde{\mathbf{M}}^{\dagger} \tilde{\mathbf{M}} \right) \mathbf{x}_0 \\
        & = \tilde{\mathbf{M}} \mathbf{o} + \mathbf{0} \\
        & = \hat{\mathbf{o}}.
    \end{aligned}
\end{equation}
Obviously, $\tilde{\mathbf{M}}^{\dagger} = \mathbf{I}$ is one of the pseudo-inverses, as
\begin{equation}
    \tilde{\mathbf{M}} \tilde{\mathbf{M}}^{\dagger} \tilde{\mathbf{M}} = \tilde{\mathbf{M}} \mathbf{I} \tilde{\mathbf{M}} = \tilde{\mathbf{M}}.
\end{equation}
Then, we can attain a succinct $\hat{\mathbf{x}}_0$ satisfying \cref{eq: consistency} as below:
\begin{equation}
    \label{eq: proper_x0}
    \hat{\mathbf{x}}_0 = \tilde{\mathbf{M}}^{\dagger} \hat{\mathbf{o}} + \left( \mathbf{I} - \tilde{\mathbf{M}}^{\dagger} \tilde{\mathbf{M}} \right) \mathbf{x}_0 = \hat{\mathbf{o}} + \left( \mathbf{I} - \tilde{\mathbf{M}} \right) \mathbf{x}_0.
\end{equation}
Roughly speaking, \cref{eq: proper_x0} tells us to fill the empty pixels of the incomplete depth map with the values from $\mathbf{x}_0$.
Following~\cite{Karras2022edm}, we use a neural network $\mathcal{Z}_{\mathbf{\theta}}$ to directly predict $\mathbf{x}_0$ instead of noises $\epsilon_t$ from $\mathbf{x}_t$ at time step $t$, which is denoted as $\mathbf{x}_{0|t}$. In practice, we apply \cref{eq: proper_x0} at each time step $t$,
\begin{equation}
    \hat{\mathbf{x}}_{0|t}  = \hat{\mathbf{o}} + \left( \mathbf{I} - \tilde{\mathbf{M}} \right) \mathbf{x}_{0|t},
\end{equation}
to generate the $\mathbf{x}_0$ that satisfies both \cref{eq: consistency,eq: realness}, as shown in \cref{fig: pipeline}.

\paragraph{\bf Intrinsics for Condition.} For the conversion between depth maps and point clouds, accurate camera intrinsics are very important. Otherwise, the resulting point clouds will be distorted. Thus, the diffusion model should be able to generate a depth map under the specific intrinsics. For this reason, we slightly modify the vanilla diffusion model, adding the intrinsics as a condition during the reverse process.
Intrinsics are compiled into a vector $\mathbf{k}=[fx, fy, cx, cy]^T$, where $fx$, $fy$, $cx$, and $cy$ denote the focal lengths and the principal points, respectively, which then processed by a 2-layer MLP and inserted into a U-Net during cross-attention as an embedding.

\subsection{Depth Correction}
\label{sec: 3.3}
\begin{figure}[t]
     \begin{minipage}[h]{0.48\linewidth}
        \centering
        \includegraphics[width=\linewidth]{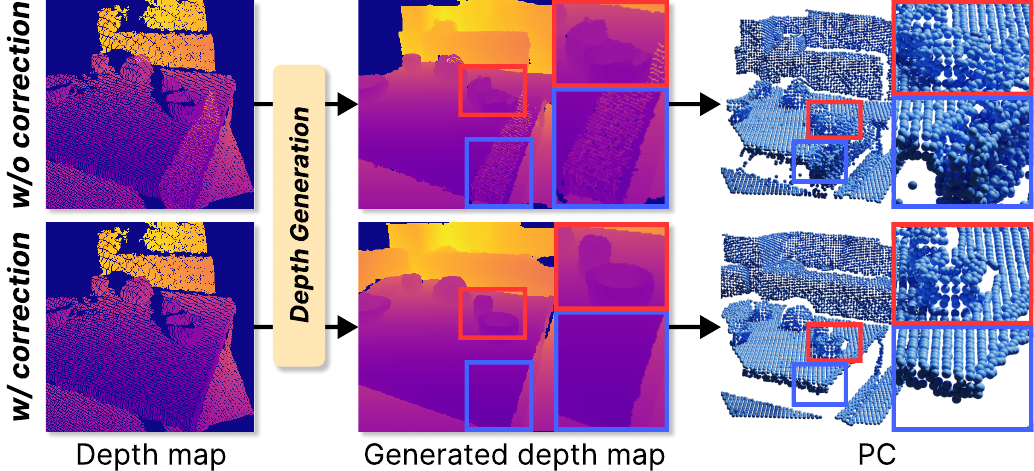} 
        \caption{Heavy artifacts will be generated if the point penetration problem is not well resolved. The unnatural pixels can be removed by depth correction to improve the generative results.} 
        \label{fig: artifacts}
    \end{minipage}
    \hfill
     \begin{minipage}[h]{0.48\linewidth}
        \centering
        \includegraphics[width=\linewidth]{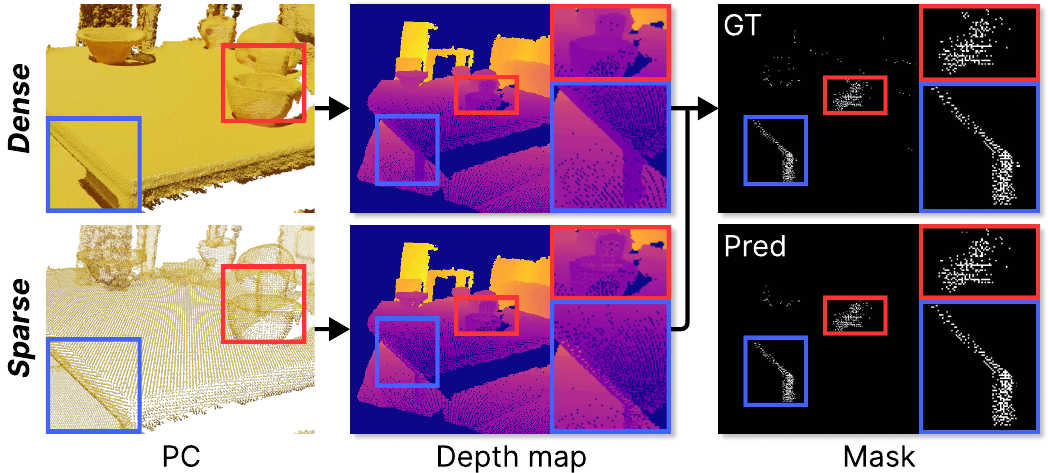} 
        \caption{Illustration of the data preparation process for training our depth correction module. For better visualization, "True" is colored in black and "False" is colored in white.} 
        \label{fig: data_preparation}
    \end{minipage}
    % \vspace{-4mm}
\end{figure}

Now, we have gained a generated depth map $\mathbf{O}$ with new content and partially overlapped with the given depth map $\mathbf{I}$. It seems the two depth maps can be simply converted to partially overlapped point clouds with the camera intrinsics and become a piece of training data.
However, the re-projected depth maps $\hat{\mathbf{O}}$ are usually unreliable, containing a lot of unnatural depth values from penetrated points, \ie, the point penetration problem. Specifically, from the novel view of re-projected depth maps, some points of the objects behind will be blended inside the objects in front due to the point cloud sparsity, which will confuse diffusion models, leading to generating heavy artifacts, as shown in \cref{fig: artifacts}. A simple solution is to densify the point cloud before re-projection by using the truncated signed distance function (TSDF) to fuse tens of depth maps for simulating occlusion under extreme density during re-projection. However, camera poses are required in this way, and the time consumption will explode while processing with so many points. To this end, we designed a depth correction module to mask out the invalid values in the depth maps directly.

\paragraph{\bf Network Architecture.} We adopt the U-Net backbone for its great performance in various tasks. Nonetheless, the vanilla U-Net faces challenges in precise awareness of  3D structures in plain depth maps. Thus, as shown in \cref{fig: pipeline}, in the \ourdepthcorretionname{}, a depth augmentation module is added before the U-Net to provide more structural information that a convolutional kernel is hard to learn. Given a re-projected depth map $\hat{\mathbf{O}}$ with invalid values, it first extracts the minimum value for each pixel from its neighbors using a min-pooling operator, yielding the minimum neighbor map, which is close to the potentially correct depth map. Then, it computes the residual value for each pixel, yielding the residual map, which is close to the expected output mask. The original incorrect depth map, the minimum neighbor map, and the residual map will be concatenated together and fed into the subsequent U-Net. Finally, the U-Net predicts the mask binarized by a specific threshold $\tau_m$, telling the correctness of a pixel.

\paragraph{\bf Data Preparation.} To prepare the training data for the \ourdepthcorretionname{}, as is shown in \cref{fig: data_preparation}, we first obtain a dense point cloud $\mathbf{P}_d \in \mathbb{R}^{N_d \times 3}$ of high density using TSDF as mentioned in \cref{sec: 3.3}, and apply voxel down-sampling with a bigger voxel size to gain the sparse version of the point cloud $\mathbf{P}_s \in \mathbb{R}^{N_s \times 3}$. Then the two point clouds $\mathbf{P}_d$ and $\mathbf{P}_s$ are re-projected to the pixel coordinates under the same pose $\mathbf{T}$, yielding two corresponding depth maps: one without the problem of point penetration, denoted by $\mathbf{I}_d$, which is re-projected from $\mathbf{P}_d$; the other having the problem of point penetration, denoted by $\mathbf{I}_s$, which is re-projected from $\mathbf{P}_s$. $\mathbf{I}_s$ is supposed to be the input image and the ground-truth mask $\mathbf{M}_{gt}$ can be computed from the difference between $\mathbf{I}_d$ and $\mathbf{I}_s$ with a threshold $\tau_{gt}$:
\begin{equation}
    \mathbf{M}_{gt}\left( i, j \right) = \left[ \left( \mathbf{I}_{s}\left( i, j \right) - \mathbf{I}_{d}\left( i, j \right) \right )^2 < \tau^2_{gt} \right]
\end{equation}
where $\mathbf{M}_{gt}\left( i, j \right)$, $\mathbf{I}_{s}\left( i, j \right)$, and $\mathbf{I}_{d}\left( i, j \right)$ denote the $i$-th row and $j$-th column of $\mathbf{M}_{gt}$, $\mathbf{I}_s$ and $\mathbf{I}_d$, respectively. $\left[ \cdot \right]$ is the Iversion bracket.

\paragraph{\bf Loss.} As the depth correction module generates a mask, which can be considered a task of binary segmentation, we simply adopt the binary cross-entropy loss between the output of U-Net and the ground-truth mask for training.

\subsection{Point Cloud Registration}
\label{sec: 3.4}
Given a source point cloud $\mathbf{P} \in \mathbb{R}^{N \times 3}$ and a target point cloud $\mathbf{Q} \in \mathbb{R}^{M \times 3}$, the objective of point cloud registration is to align the source to the target by estimating the rotation $\mathbf{R} \in  SO(3)$ and translation $\mathbf{t} \in \mathbb{R}^3$ between them.

As shown in \cref{fig: pipeline}, recent pipelines for point cloud registration mainly consist of three steps: (i) extracting point-wise or path-wise features with a feature descriptor, (ii) feature matching to find correspondences, and (iii) estimating transformations from correspondences using a traditional or deep robust estimator.

We validate the effectiveness of our approach by training previous state-of-the-art registration models~\cite{huang2020predator, yu2021cofinet, qin2022geometric} using our generated dataset as supplementary data and evaluate them on real-world benchmarks.

\section{Experiments}
\subsection{Datasets and Evaluation Metrics.}
\noindent{\bf 3DMatch~\cite{zeng20173dmatch}:} 3DMatch is a widely-used indoor RGB-D dataset containing $62$ scenes among which $46$/$8$/$8$ scenes are used for training/validation/test, respectively. 
Following previous works~\cite{zeng20173dmatch}, we employ depth maps from the training set to train all our modules.
For evaluation, following PREDATOR~\cite{huang2020predator}, we split the test set into 3DMatch and 3DLoMatch whose point cloud pairs have $>30\%$ and $10\%-30\%$ overlap. 

\noindent{\bf ETH~\cite{pomerleau2012challenging}:} ETH is an outdoor dataset used only for testing; it contains 713 pairs made up of 132 point clouds from 4 scenes collected by a laser scanner. It is extremely challenging due to the luxuriant trees and small facilities. 

\noindent{\bf Evaluation metrics:}
Following~\cite{bai2020d3feat, huang2020predator, qin2022geometric, yew2022regtr, wang2023roreg}, we use the following metrics to evaluate the performance in point cloud registration: (i) Registration Recall (RR), the fraction of successfully registered point cloud pairs whose transformation error RMSE $< 0.2 \mathrm{m}$/$0.5 \mathrm{m}$ for 3D(Lo)Match/ETH; (ii) Inlier Ratio (IR), the fraction of inlier correspondences whose residuals $< 0.1 \mathrm{m}$/$0.2 \mathrm{m}$ for 3D(Lo)Match/ETH among all hypothesized correspondences; (iii) Feature Matching Recall (FMR), the fraction of point cloud pairs whose IR $> 5\%$; (iv) the median of the average Relative Rotation Error (RRE); and (v) the median of the average Relative Translation Error (RTE) for the successfully registered pairs whose RMSE $< 0.2 \mathrm{m}$/$0.5 \mathrm{m}$ for 3D(Lo)Match/ETH.

\subsection{Implementation Details.}
We train our \ourdiffusionname{} using $32$ GeForce RTX 2080 Tis with a batch size of $128$ for $2,000k$ iterations.
For the forward process, we set $T=1000$ and use the sigmoid schedule~\cite{jabri2022scalable} for $\beta_t$. For the reverse process, following~\cite{song2020denoising}, we set $T=250$ to accelerate the generation process. 
To train our \ourdepthcorretionname{}, we use $8$ GeForce RTX 2080 Tis with a batch size of $32$ for $50$ epochs. We use the Adam optimizer with an initial learning rate of $1e^{-4}$, which is decayed by $0.95$ for each epoch.
To evaluate the effectiveness of our generative dataset, we choose the previous state-of-the-art methods~\cite{huang2020predator, yu2021cofinet, qin2022geometric} as baselines and train them using the generative dataset as an additional dataset, following all the training and testing protocols but increasing the total training epochs.
For a fair comparison, we generate $160k$ pairs in total for training in \cref{tab: comparison_sira,tab: boosting_indoor,tab: boosting_outdoor}, 
the quantity of which is close to the additional data that used in SIRA-PCR~\cite{chen2023sira}. We only use the first $20k$ pairs in the ablation studies in \cref{tab:ablation_depth_correction}. For the results obtained by RANSAC, we follow~\cite{bai2020d3feat, huang2020predator, qin2022geometric, wang2023roreg}, running $50k$ iterations to estimate the transformation and setting the inlier distance threshold to $0.05\mathrm{m}/0.4\mathrm{m}$ for 3D(Lo)Match/ETH, respectively. Please refer to our supp. material for other details.

\subsection{Evaluation on Indoor Benchmark}
\begin{table}[t]
    \setlength{\tabcolsep}{5.0pt}
    \centering
    \caption{Comparison of the generated datasets. The same model GeoTrans~\cite{qin2022geometric} is learned on 3DMatch without and with additional rendered (+ SIRA~\cite{chen2023sira}) or our generated data (+ Ours). The best and second-best results are marked in \textbf{bold} and \underline{underlined}.
    }
    \resizebox{0.8\linewidth}{!}{
        \begin{tabular}{l|l|ccc|ccc}
        \toprule
         \multirow{2}{*}{Methods} & \multirow{2}{*}{Data}& \multicolumn{3}{c|}{3DMatch} & \multicolumn{3}{c}{3DLoMatch} \\
        &&  FMR $\uparrow$ & IR $\uparrow$ & RR $\uparrow$ & FMR $\uparrow$ & IR $\uparrow$ & RR $\uparrow$ \\
         \midrule
        GeoTrans~\cite{qin2022geometric} &3DMatch& \underline{97.7} & 70.3 & 91.5 & 88.1 & 43.3 & 74.0 \\
        + SIRA~\cite{chen2023sira} &+ Rendered data& \textbf{98.7} &\underline{69.8} & \textbf{94.1} & \underline{88.5} & \underline{42.9} & \underline{76.6} \\
        + Ours &+ Generated data& \textbf{98.7} & \textbf{71.9} & \underline{93.3} & \textbf{89.4} & \textbf{45.6} & \textbf{77.2} \\
        \bottomrule
    \end{tabular}}
    \label{tab: comparison_sira}
\end{table}
To assess the efficacy of our proposed method for realistic data generation, we compare our method with the baseline method GeoTrans~\cite{qin2022geometric} trained solely on the 3DMatch dataset and the state-of-the-art method SIRA-PCR~\cite{chen2023sira}, where additional $160k$ synthetic training pairs are used to train GeoTrans.
Note that we only use depth maps from 3DMatch for data generation, so no additional data sources are used in our method. As can be seen from \cref{tab: comparison_sira}, on the test set of 3DMatch, our method improves the performance of the baseline method GeoTrans by $1.0$/$1.6$/$1.8$ percentage points (pp) in FMR/IR/RR. Though additional synthetic data source is used in SIRA-PCR, our method can achieve comparable results and even improve the IR metric by $2.1$ compared with SIRA-PCR. More impressively, for more complex scenarios with lower overlaps in 3DLoMatch benchmark, our method enhances the baseline performance by $1.3$/$2.3$/$3.2$ pp in FMR/IR/RR and surpasses SIRA-PCR~\cite{chen2023sira} in the metrics of FMR/IR/RR by $0.9$/$2.7$/$0.6$ pp, consistently. This demonstrates the effectiveness of our method in generating realistic 3D point cloud registration data.
\begin{table*}[t]
\setlength{\tabcolsep}{2.3pt}
\scriptsize
    \centering
    \caption{Evaluation results on 3DMatch and 3DLoMatch. The results boosted from those officially reported are marked in \textbf{bold} for a better comparison. *: the results are reproduced using the official codes without any configuration changes.}
    \resizebox{\linewidth}{!}{
    \begin{tabular}{l|l|ccccc|ccccc|ccccc}
        \toprule
        & \multirow{2}{*}{\# Samples} & \multicolumn{5}{c|}{Feature Matching Recall (\%) $\uparrow$} & \multicolumn{5}{c|}{Inlier Ratio (\%) $\uparrow$} & \multicolumn{5}{c}{Registration Recall (\%) $\uparrow$} \\
        & & 5000 & 2500 & 1000 & 500 & 250 & 5000 & 2500 & 1000 & 500 & 250  & 5000 & 2500 & 1000 & 500 & 250 \\
        \midrule
        \multirow{13}{*}{\rotatebox{90}{3DMatch}}
        & PerfectMatch~\cite{gojcic2019perfect} & 95.0 & 94.3 & 92.9 & 90.1 & 82.9           & 36.0 & 32.5 & 26.4 & 21.5 & 16.4           & 78.4 & 76.2 & 71.4 & 67.6 & 50.8 \\
        & FCGF~\cite{choy2019fully} & 97.4 & 97.3 & 97.0 & 96.7 & 96.6            & 56.8 & 54.1 & 48.7 & 42.5 & 34.1           & 85.1 & 84.7 & 83.3 & 81.6 & 71.4  \\
        & D3Feat~\cite{bai2020d3feat} & 95.6 & 95.4 & 94.5 & 94.1 & 93.1           & 39.0 & 38.8 & 40.4 & 41.5 & 41.8            & 81.6 & 84.5 & 83.4 & 82.4 & 77.9 \\
        & SpinNet~\cite{ao2021spinnet} & 97.6 & 97.2 & 96.8 & 95.5 & 94.3           & 47.5 & 44.7 & 39.4 & 33.9 & 27.6            & 88.6 & 86.6 & 85.5 & 83.5 & 70.2 \\
        & YOHO~\cite{wang2022you} & 98.2 & 97.6 & 97.5 & 97.7 & 96.0           & 64.4 & 60.7 & 55.7 & 46.4 & 41.2           & 90.8 & 90.3 & 89.1 & 88.6 & 84.5 \\
        & OIF-PCR~\cite{yangone} & 98.1 & 98.1 & 97.9 & 98.4 & 98.4           & 62.3 & 65.2 & 66.8 & 67.1 & 67.5            & 92.4 & 91.9 & 91.8 & 92.1 & 91.2 \\
        & RoReg~\cite{wang2023roreg} & 98.2 & 97.9 & 98.2 & 97.8 & 97.2           & 81.6 & 80.2 & 75.1 & 74.1 & 75.2           & 92.9 & 93.2 & 92.7 & 93.3 & 91.2 \\
        & PEAL~\cite{yu2023peal} & 99.0 & 99.0 & 99.1 & 99.1 & 98.8          & 72.4 & 79.1 & 84.1 & 86.1 & 87.3           & 94.6 & 93.7 & 93.7 & 93.9 & 93.4 \\
        \cmidrule{2-17}
        & PREDATOR~\cite{huang2020predator} & 96.6 & 96.6 & 96.5 & 96.3 & 96.5           & 58.0 & 58.4 & 57.1 & 54.1 & 49.3           & 89.0 & 89.9 & 90.6 & 88.5 & 86.6 \\
        & PREDATOR* & 96.9 & 96.6 & 96.6 & 96.5 & 96.8           & 57.8 & 58.1 & 56.9 & 54.3 & 50.2           & 88.5 & 89.3 & 88.0 & 88.4 & 85.7 \\
        \rowcolor{highlight}
        \cellcolor{white} & PREDATOR + Ours & \textbf{98.3} & \textbf{98.0} & \textbf{98.3} & \textbf{98.3} & \textbf{98.1}           & \textbf{64.8} & \textbf{65.1} & \textbf{63.7} & \textbf{60.8} & \textbf{55.6}           & \textbf{90.8} & \textbf{90.5} & \textbf{90.7} & \textbf{90.1} & \textbf{89.6} \\
        \cmidrule{2-17}
        & CoFiNet~\cite{yu2021cofinet}  & 98.1 & 98.3 & 98.1 & 98.2 & 98.3           & 49.8 & 51.2 & 51.9 & 52.2 & 52.2           & 89.3 & 88.9 & 88.4 & 87.4 & 87.0  \\
        & CoFiNet*  & 97.6 & 97.5 & 97.7 & 97.6 & 97.5           & 43.2 & 44.9 & 45.7 & 46.0 & 46.2          & 88.2 & 88.8 & 88.2 & 88.5 & 86.7 \\
        \rowcolor{highlight}
        \cellcolor{white} & CoFiNet + Ours & 98.1 & 98.1 & \textbf{98.2} & 98.2 & 98.1           & \textbf{49.9}  & 51.0 & 51.7 & 51.8 & 51.9           & \textbf{90.6} & \textbf{90.2} & \textbf{90.1} &\textbf{ 90.1} & \textbf{89.1}\\
        \cmidrule{2-17}
        & GeoTrans~\cite{qin2022geometric} & 97.9 & 97.9 & 97.9 & 97.9 & 97.6          & 71.9 & 75.2 & 76.0 & 82.2 & 85.1           & 92.0 & 91.8 & 91.8 & 91.4 & 91.2 \\
        & GeoTrans* & 98.2 & 98.2 & 98.0 & 98.0 & 98.0          & 72.4 & 77.5 & 82.2 & 84.1 & 85.3           & 91.4 & 91.3 & 90.7 & 90.7 & 90.7 \\
        \rowcolor{highlight}
        \cellcolor{white} & GeoTrans + Ours & \textbf{98.7} & \textbf{98.7} & \textbf{98.6} & \textbf{98.5} & \textbf{98.2}           & \textbf{73.8} & \textbf{80.5} & \textbf{84.9} & \textbf{86.7} & \textbf{87.8}           & \textbf{92.7} & \textbf{92.5} & \textbf{92.2} & \textbf{91.6} & \textbf{91.3} \\
        \midrule
        \multirow{13}{*}{\rotatebox{90}{3DLoMatch}}
        & PerfectMatch~\cite{gojcic2019perfect} & 63.6 & 61.7 & 53.6 & 45.2 & 34.2            & 11.4 & 10.1 & 8.0 & 6.4 & 4.8           & 33.0 & 29.0 & 23.3 & 17.0 & 11.0 \\
        & FCGF~\cite{choy2019fully} & 76.6 & 75.4 & 74.2 & 71.7 & 67.3           & 21.4 & 20.0 & 17.2 & 14.8 & 11.6           & 40.1 & 41.7 & 38.2 & 35.4 & 26.8 \\
        & D3Feat~\cite{bai2020d3feat} & 67.3 & 66.7 & 67.0 & 66.7 & 66.5            & 13.2 & 13.1 & 14.0 & 14.6 & 15.0           & 37.2 & 42.7 & 46.9 & 43.8 & 39.1 \\
        & SpinNet~\cite{ao2021spinnet} & 75.3 & 74.9 & 72.5 & 70.0 & 63.6           & 20.5 & 19.0 & 16.3 & 13.8 & 11.1           & 59.8 & 54.9 & 48.3 & 39.8 & 26.8 \\
        & YOHO~\cite{wang2022you} & 79.4 & 78.1 & 76.3 & 73.8 & 69.1           & 25.9 & 23.3 & 22.6 & 18.2 & 15.0            & 65.2 & 65.5 & 63.2 & 56.5 & 48.0  \\
        & OIF-PCR~\cite{yangone} & 84.6 & 85.2 & 85.5 & 86.6 & 87.0           & 27.5 & 30.0 & 31.2 & 32.6 & 33.1           & 76.1 & 75.4 & 75.1 & 74.4 & 73.6 \\
        & RoReg~\cite{wang2023roreg} & 82.1 & 82.1 & 81.7 & 81.6 & 80.2           & 39.6 & 39.6 & 34.0 & 31.9 & 34.5           & 70.3 & 71.2 & 69.5 & 67.9 & 64.3\\
        & PEAL~\cite{yu2023peal} & 91.7 & 92.4 & 92.5 & 92.9 & 92.7          & 45.0 & 50.9 & 57.4 & 60.3 & 62.2          & 81.7 & 81.2 & 80.8 & 80.4 & 80.1 \\
        \cmidrule{2-17}
       & PREDATOR~\cite{huang2020predator} & 78.6 & 77.4 & 76.3 & 75.7 & 75.3           & 26.7 & 28.1 & 28.3 & 27.5 & 25.8           & 59.8 & 61.2 & 62.4 & 60.8 & 58.1 \\
        & PREDATOR* & 73.4 & 74.4 & 75.3 & 74.8 & 74.4           & 22.6 & 23.9 & 24.7 & 23.8 & 22.4           & 56.2 & 57.7 & 58.7 & 57.4 & 53.7 \\
        \rowcolor{highlight}
        \cellcolor{white}& PREDATOR + Ours & \textbf{82.4} & \textbf{83.0} & \textbf{82.4} & \textbf{82.7} & \textbf{82.2}           & \textbf{33.1} & \textbf{34.6} & \textbf{35.1} & \textbf{34.4} & \textbf{32.2}           & \textbf{63.0} & \textbf{65.1} & \textbf{67.2} & \textbf{66.0} & \textbf{63.4} \\
        \cmidrule{2-17}
        & CoFiNet~\cite{yu2021cofinet} & 83.1 & 83.5 & 83.3 & 83.1 & 82.6           & 24.4 & 25.9 & 26.7 & 26.8 & 26.9          & 67.5 & 66.2 & 64.2 & 63.1 & 61.0 \\
        & CoFiNet* & 79.7 & 80.5 & 80.5 & 79.9 & 80.2           & 18.8 & 20.2 & 21.0 & 21.3 & 21.2        & 61.4 & 61.8 & 61.3 & 60.9 & 59.3 \\
        \rowcolor{highlight}
        \cellcolor{white}& CoFiNet + Ours & \textbf{85.1} & \textbf{85.1} & \textbf{85.1} & \textbf{85.0} & \textbf{84.2}             & \textbf{25.5} & \textbf{26.7} & \textbf{27.4} & \textbf{27.5} & \textbf{27.7}           & 67.5 & \textbf{67.0} & \textbf{67.8} & \textbf{66.8} & \textbf{64.4} \\
        \cmidrule{2-17}
        & GeoTrans~\cite{qin2022geometric}  & 88.3 & 88.6 & 88.8 & 88.6 & 88.3          & 43.5 & 45.3 & 46.2 & 52.9 & 57.7            & 75.0 & 74.8 & 74.2 & 74.1 & 73.5  \\
        & GeoTrans* & 87.2 & 87.2 & 87.4 & 87.4 & 87.1          & 43.7 & 49.1 & 55.4 & 58.0 & 59.5           & 70.4 & 70.8 & 70.5 & 70.0 & 69.5 \\
        \rowcolor{highlight}
        \cellcolor{white}& GeoTrans + Ours & \textbf{89.4} & \textbf{89.6} & \textbf{89.5} & \textbf{89.4} & \textbf{88.7}           & \textbf{45.9} & \textbf{52.0} & \textbf{58.2} & \textbf{60.8} & \textbf{62.6}           & 74.8 & 74.2 & 73.8 & 74.0 & 72.6 \\
        \bottomrule
    \end{tabular}}
    \label{tab: boosting_indoor}
\end{table*}

\begin{figure*}[t]
    \centering
    \includegraphics[width=\linewidth]{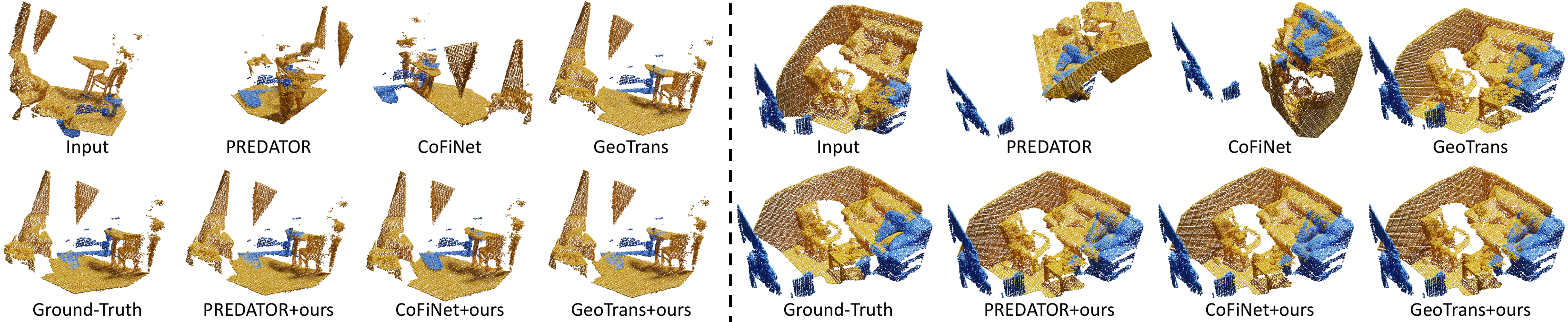}
    \caption{Qualitative results with baselines~\cite{huang2020predator, yu2021cofinet, qin2022geometric} on 3DMatch (left)  and 3DLoMatch (right)}
    \label{fig: qualitative_indoor}
\end{figure*}

To further validate the robustness of our method, we learn different deep networks such as PREDATOR~\cite{huang2020predator}, CoFiNet~\cite{yu2021cofinet}, and GeoTrans~\cite{qin2022geometric}. Evaluations are conducted under varying numbers of sampled correspondences with RANSAC. As detailed in \cref{tab: boosting_indoor}, our method significantly enhances PREDATOR across all metrics, with improvements of $1.4$/$6.3$/$0.1$ pp at least in FMR/IR/RR on 3DMatch, and $3.8$/$6.4$/$3.2$ pp at least on 3DLoMatch. CoFiNet also benefits, particularly in RR, with increases of $1.3\text{–}2.7$ pp and $0.0\text{–}3.7$ pp on 3DMatch and 3DLoMatch, respectively. While GeoTrans shows an overall enhancement on 3DMatch, a slight decrease in RR is observed on 3DLoMatch; however, FMR and IR still improve by $0.4\text{–}1.1$ pp and $2.4\text{–}12.0$ pp, respectively. This demonstrates that our approach can consistently enhance the performance of learning-based methods across various metrics on the indoor 3DMatch benchmark. \Cref{fig: qualitative_indoor} provides qualitative comparisons, highlighting the robustness and versatility of our method in challenging scenarios with partial geometric structures and low overlap ratios.

\subsection{Evaluation on Outdoor Benchmark}
\begin{table*}[t]
\setlength{\tabcolsep}{2.3pt}
\scriptsize
    \centering
    \caption{Evaluation results on ETH. \textemdash: results with different samples are not applicable for LGR since it uses all correspondences. The boosted results are marked in \textbf{bold} for a better comparison.}
    \resizebox{\linewidth}{!}{
    \begin{tabular}{l|ccccc|ccccc|ccccc}
        \toprule
        \multirow{2}{*}{\# Samples} & \multicolumn{5}{c|}{Feature Matching Recall (\%) $\uparrow$} & \multicolumn{5}{c|}{Inlier Ratio (\%) $\uparrow$} & \multicolumn{5}{c}{Registration Recall (\%) $\uparrow$} \\
        & 5000 & 2500 & 1000 & 500 & 250 & 5000 & 2500 & 1000 & 500 & 250  & 5000 & 2500 & 1000 & 500 & 250 \\
        \midrule
        PerfectMatch~\cite{gojcic2019perfect} & 95.6 & 94.3 & 80.5 & 69.1 & 51.4 & 19.7 & 16.7 & 12.4 & 9.3 & 6.6 & 81.4 & 73.5 & 59.3 & 46.5 & 35.0 \\
        D3Feat~\cite{bai2020d3feat} & 63.3 & 71.0 & 69.7 & 67.9 & 60.5 & 12.5 & 13.2 & 13.6 & 13.5 & 11.8 & 59.1 & 50.4 & 49.7 & 44.6 & 29.1 \\
        FCGF~\cite{choy2019fully} & 41.1 & 38.4 & 32.3 & 24.6 & 15.9 & 5.8 & 5.3 & 4.4 & 3.5 & 2.8 & 42.1 & 36.1 & 29.5 & 26.3 & 18.9 \\
        SpinNet~\cite{ao2021spinnet} & 99.4 & 99.1 & 94.6 & 87.2 & 66.6 & 23.2 & 20.4 & 15.7 & 11.9 & 8.6 & 96.0 & 91.1 & 81.9 & 71.5 & 54.3 \\
        RoReg~\cite{wang2023roreg} & 96.5 & 95.6 & 93.1 & 92.0 & 84.1 & 28.4 & 25.4 & 21.7 & 20.8 & 17.5 & 97.1 & 97.1 & 95.7 & 92.3 & 84.3 \\
        \midrule
        PREDATOR~\cite{huang2020predator} & 65.6 & 64.5 & 59.6 & 52.0 & 40.5 & 11.1 & 10.3 & 8.5 & 6.8 & 5.1 & 74.7 & 72.9 & 67.7 & 60.3 & 51.7 \\
        \rowcolor{highlight}
        PREDATOR + Ours & \textbf{76.9} & \textbf{78.0} & \textbf{73.3 }& \textbf{70.8} & \textbf{67.1} & \textbf{15.1} & \textbf{14.6} & \textbf{13.2} & \textbf{12.0} & \textbf{10.5} & \textbf{84.1} & \textbf{81.1} & \textbf{75.1} & \textbf{65.6} & \textbf{51.9} \\
         \midrule
        CoFiNet~\cite{yu2021cofinet} & 82.5 & 83.7 & 81.9 & 81.1 & 79.9 & 9.6 & 9.8 & 9.9 & 9.9 & 9.8 & 83.9 & 82.7 & 81.9 & 77.4 & 68.8 \\
        \rowcolor{highlight}
        CoFiNet + Ours & \textbf{89.3} & \textbf{90.1} & \textbf{89.7} & \textbf{88.6} & \textbf{89.2} & \textbf{10.0} & \textbf{10.2} & \textbf{10.3} & \textbf{10.3} & \textbf{10.2} & \textbf{88.0} & \textbf{88.6} & \textbf{88.1} & \textbf{85.3} & \textbf{76.7} \\
        \midrule
        GeoTrans~\cite{qin2022geometric} & \multicolumn{2}{c}{\textemdash} & 59.9 & \multicolumn{2}{c|}{\textemdash} & \multicolumn{2}{c}{\textemdash} & 6.7 & \multicolumn{2}{c|}{\textemdash} & \multicolumn{2}{c}{\textemdash} & 85.5 & \multicolumn{2}{c}{\textemdash} \\
        \rowcolor{highlight}
        GeoTrans + Ours & \multicolumn{2}{c}{\textemdash} & 56.1 & \multicolumn{2}{c|}{\textemdash} & \multicolumn{2}{c}{\textemdash} & 6.1 & \multicolumn{2}{c|}{\textemdash} & \multicolumn{2}{c}{\textemdash} & \textbf{85.8} & \multicolumn{2}{c}{\textemdash} \\
        \bottomrule
    \end{tabular}}
    \label{tab: boosting_outdoor}
\end{table*}
\begin{figure*}[t]
    \centering
    \includegraphics[width=\linewidth]{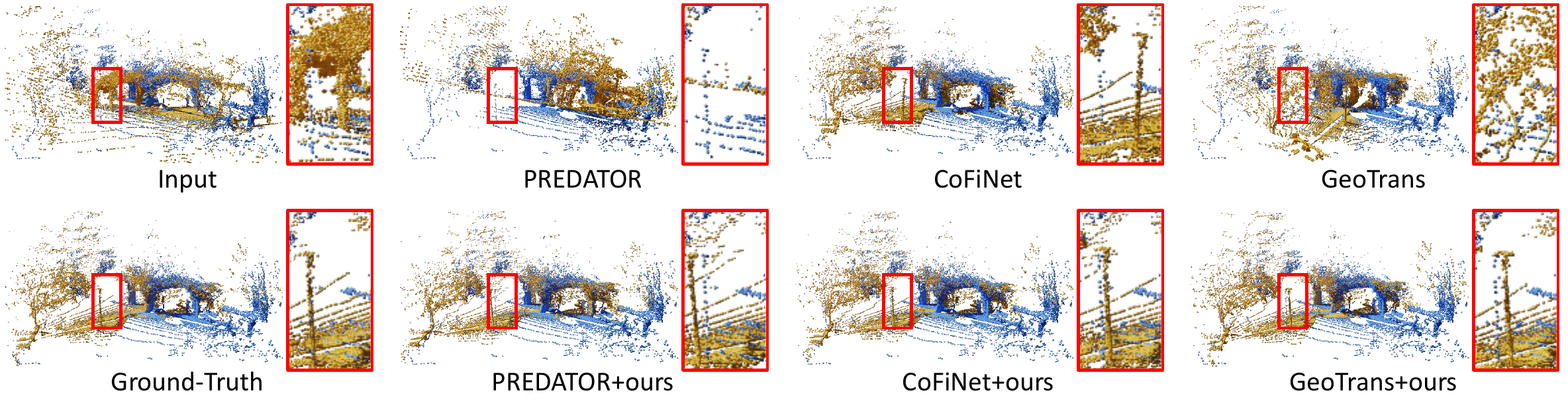}
    \caption{Qualitative comparison with state-of-the-art methods~\cite{huang2020predator, yu2021cofinet, qin2022geometric} on the ETH benchmark}
    \label{fig: qualitative_outdoor}
\end{figure*}
In order to assess the generalization capacity of our method, following recent studies~\cite{wang2023roreg, chen2023sira}, we directly evaluate our learned models on the ETH benchmark without any fine-tuning. Results are shown in \cref{tab: boosting_outdoor}, where both PREDATOR and CoFiNet benefit from the inclusion of our generated data when applied to unseen outdoor datasets. More specifically, PREDATOR shows notable improvements of $11.3\text{–}26.6$ pp in FMR, $4.0\text{–}5.4$ pp in IR, and $0.2\text{–}9.4$ pp in RR. Similarly, CoFiNet also experiences enhancements due to our method, with gains of $6.8\text{–}9.3$ pp in FMR, a consistent $0.4$ in IR, and $4.1\text{–}7.9$ pp in RR. 
The qualitative comparisons are depicted in \cref{fig: qualitative_outdoor}, where our approach notably enhances the performance of all baseline methods, particularly in challenging scenarios characterized by complex structures and extreme noise, which are typical in unseen outdoor environments. This demonstrates the robustness and generalization ability of our method.

\subsection{Ablation Study}
\label{sec: 4.5}
Our ablation studies adopt GeoTrans~\cite{qin2022geometric} with its LGR as the baseline for stable results.

\begin{table}[t]
     \begin{minipage}[h]{0.48\linewidth}
        \centering
        \caption{Ablation studies of the depth generation module}
        \resizebox{\linewidth}{!}{
        \begin{tabular}{l|ccc|ccc}
            \toprule
             \multirow{2}{*}{Methods} & \multicolumn{3}{c|}{3DMatch}& \multicolumn{3}{c}{3DLoMatch} \\
             & FMR $\uparrow$ & IR $\uparrow$ & RR $\uparrow$ & FMR $\uparrow$ & IR $\uparrow$ & RR $\uparrow$ \\
             \midrule
             (a) baseline & 97.7 & 70.3 & \underline{91.5} & 88.1 & \underline{43.3} & \underline{74.0}\\
             (b) re-project. only & \underline{98.2} & 69.6  & 91.2  & \underline{88.7} & 41.1 & 72.1 \\
             (c) w/o depth generat. & \textbf{98.3} & \underline{70.4} & \underline{91.5} & 87.4 & 42.4  & 73.9 \\
             (d) w/ depth generat. & 98.0 & \textbf{71.6} & \textbf{91.9} & \textbf{89.4} & \textbf{44.9} & \textbf{76.5} \\
    
             \bottomrule
        \end{tabular}}
        \label{tab:ablation_no_generation}
    \end{minipage}
    \hfill
     \begin{minipage}[h]{0.48\linewidth}
        \centering
        \caption{Ablation studies of different dataset generation processes}
        \resizebox{\linewidth}{!}{
        \begin{tabular}{l|ccc|ccc}
            \toprule
             \multirow{2}{*}{Methods} & \multicolumn{3}{c|}{3DMatch}& \multicolumn{3}{c}{3DLoMatch} \\
             & FMR $\uparrow$ & IR $\uparrow$ & RR $\uparrow$ & FMR $\uparrow$ & IR $\uparrow$ & RR $\uparrow$ \\
             \midrule
             (a) baseline & 97.7 & 70.3 & 91.5 & \underline{88.1} & \underline{43.3} & \underline{74.0}\\
             (b) unconditional & 97.6 & \underline{70.8} & \underline{91.8} & 86.9 & 42.3 & 73.2 \\
             (c) w/o depth correct. & \textbf{98.5} & 70.6 & 91.3 & 87.8 & 42.6 & 73.5 \\
             (d) w/ depth correct. & \underline{98.0} & \textbf{71.6} & \textbf{91.9} & \textbf{89.4} & \textbf{44.9} & \textbf{76.5} \\
             \bottomrule
        \end{tabular}}
        \label{tab:ablation_depth_correction}
    \end{minipage}
\end{table}
\paragraph{\bf Depth Generation Module}
is a necessary part of our pipeline, which introduces new information to augment the dataset by generating new content in the incomplete depth maps after re-projection. As Rows~(b-c) in \cref{tab:ablation_no_generation} shows, the performance degrades when removing the depth generation module, since no new information is introduced.
Plus, the Row~(b) of \cref{tab:ablation_no_generation} shows that the baseline will not be boosted if the depth correction module is removed as well and it remains re-projection only in the pipeline, because the point clouds in pairs generated in this way have many points exactly in the same positions, leading to overfitting when contrastive learning is applied.

\paragraph{\bf Unconditional Generation.}
It is possible to unconditionally generate the first depth map in our pipeline by diffusion models~\cite{ho2020denoising, song2020denoising}. As such, the point cloud registration training data can be generated without the need of real depth maps. However, we find the quality of data generated in this way unstable and unreliable, due to the unpredictable process of unconditional generation impairing the performance of models trained on it.  As shown in \cref{tab:ablation_depth_correction}, the performance of the model trained on the data generated without real depths has little improvement on 3DMatch and even drops sharply on 3DLoMatch.

\paragraph{\bf Depth Correction Module.}
As mentioned in Sec.~\ref{sec: 3.3}, the \ourdepthcorretionname{} is crucial in our pipeline to solve the point penetration problem.
In the Row~(c) of \cref{tab:ablation_depth_correction}, when the \ourdepthcorretionname{} is disabled, the performance of the learned model drops and is even lower than the baseline model learned without any generated data. This demonstrates the design of our \ourdepthcorretionname{} for dealing with the point penetration problem and ensuring the realism of the generated data. 

\begin{table}[t]
    \centering
    \caption{Ablation studies of depth augmentation}
    \setlength{\tabcolsep}{3pt}
    \resizebox{0.6\linewidth}{!}{
    \begin{tabular}{l|ccc|ccc}
        \toprule
          \multirow{2}{*}{Methods} &  \multicolumn{3}{c|}{$\tau_m = 0.5$}&  \multicolumn{3}{c}{$\tau_m = 0.99$}\\
          &  mIoU $\uparrow$ &  PAcc $\uparrow$ &  FP $\downarrow$  &  mIoU $\uparrow$ &  PAcc $\uparrow$ &  FP $\downarrow$ \\
         \midrule
         handcrafted~\cite{abs-2106-10859} & 59.8 & \underline{96.4} & 362.90 & 59.8 & \underline{96.4} & 362.90 \\
         w/o depth augment. & \underline{84.6} & \textbf{99.8} & \underline{75.84}  & \underline{74.8} & \textbf{99.2} & \underline{3.17} \\
         w/ depth augment. & \textbf{84.8} & \textbf{99.8} & \textbf{73.81} & \textbf{75.3} & \textbf{99.2} & \textbf{2.27} \\
         \bottomrule
    \end{tabular}}
    \label{tab:ablation_depth_augmentation}
\end{table}
\begin{figure}[t]
    \centering
    \includegraphics[width=\linewidth]{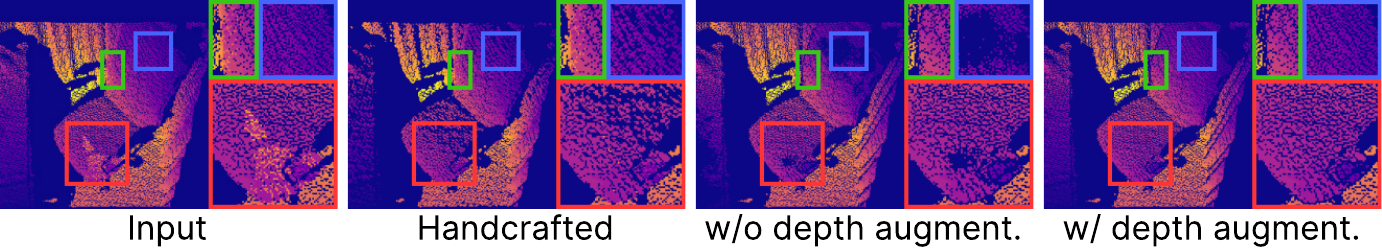}
    \caption{Comparison with different settings for depth correction}
    \label{fig: qualitative_depth_aug}
\end{figure}
\paragraph{\bf Depth Augmentation Module.}
Since depth correction is similar to image segmentation, we also use mean Intersection over Union (mIoU) and Pixel Accuracy (PAcc) to evaluate our depth correction module. In~\cite{abs-2106-10859}, a handcrafted median filter is proposed to handle the point penetration problem but fails in adapting to complex cases. As shown in \cref{tab:ablation_depth_augmentation}, the handcrafted filter achieves satisfying PAcc yet ordinary mIoU, while the \ourdepthcorretionname{} achieves both high PAcc and mIoU. However, as shown in the qualitative comparison in \cref{fig: qualitative_depth_aug}, some challenging cases cannot be handled by the vanilla U-Net. The \ourdepthaugmentationname{} is designed to enhance the perception and awareness of the \ourdepthcorretionname{} to identify the invalid penetrated points by providing more information for the subsequent processes. We focus on eliminating the unnatural depth values generated by penetrated points and thus use the False Positive (FP) to measure the ability to exclude wrong depth values, showing the average number of invalid pixels after depth correction. As shown in \cref{tab:ablation_depth_augmentation}, our depth correction module always achieves better mIoU and FP under different thresholds.

\begin{table}[ht]
    \setlength{\tabcolsep}{5.0pt}
    \centering
    \caption{Evaluation results on 3D(Lo)Match with different numbers of additional pairs. The baseline results with 0 extra pairs are reproduced using the official codes without any configuration changes. 
    The best and second-best results for each method are marked in \textbf{bold} and \underline{underlined}.
    }
    \resizebox{0.7\linewidth}{!}{
        \begin{tabular}{l|l|ccc|ccc}
        \toprule
         \multirow{2}{*}{Methods} & \multirow{2}{*}{\# Pairs}& \multicolumn{3}{c|}{3DMatch} & \multicolumn{3}{c}{3DLoMatch} \\
        &&  FMR $\uparrow$ & IR $\uparrow$ & RR $\uparrow$ & FMR $\uparrow$ & IR $\uparrow$ & RR $\uparrow$ \\
        \midrule
        \multirow{4}{*}{PREDATOR~\cite{huang2020predator}} & 0 & 96.6 & 56.9 & 88.0 & 75.3 & 24.7 & 58.7\\
        & 20k & 97.2 & 59.4 & \underline{91.1} & 80.4 & 30.5 & 65.1\\
        & 40k & \underline{98.2} & \underline{60.8} & \textbf{91.3} & \underline{81.6} & \underline{32.1} & \textbf{67.3}\\
        & 160k & \textbf{98.3} & \textbf{63.7} & 90.7 & \textbf{82.4} & \textbf{35.1} & \underline{67.2}\\
        \midrule
        \multirow{4}{*}{CoFiNet~\cite{yu2021cofinet}} & 0 & 97.6 & 43.2 & 88.2 & 79.7 & 18.8 & 61.4\\
        & 20k & \underline{97.9} & 46.5 & \underline{90.1} & 82.4 & 21.8 & 64.3\\
        & 40k & \underline{97,9} & \underline{48.2} & \underline{90.1} & \underline{83.0} & \underline{23.2} & \underline{64.7}\\
        & 160k & \textbf{98.1} & \textbf{49.9} & \textbf{90.6} & \textbf{85.1} & \textbf{25.5} & \textbf{67.5}\\
         \midrule
        \multirow{4}{*}{GeoTrans~\cite{qin2022geometric}} & 0 & 97.8 & 69.2 & 91.4 & 88.0 & 42.3 & 73.5 \\
        & 20k & 98.0 & 71.6 & 91.9 & \textbf{89.4} & \underline{44.9} & \underline{76.5} \\
        & 40k & \underline{98.1} & \textbf{72.1} & \underline{92.6} & \underline{88.6} & \textbf{45.6} & 75.9 \\
        & 160k & \textbf{98.7} & \underline{71.9} & \textbf{93.3} & \textbf{89.4} & \textbf{45.6} & \textbf{77.2} \\
        \bottomrule
    \end{tabular}}
    \label{tab: boosting_wrt_data_quantity_lgr}
\end{table}
\paragraph{\bf Data Quantity.}
We report the performance of different learning-based methods boosted under various quantities of our generative data. \Cref{tab: boosting_wrt_data_quantity_lgr} shows the performance of the baselines is boosted consistently when the number of additional generative pairs grows. As the baselines can still be largely boosted by adding more informative data, it shows the saturation point is much larger than the quantity of commonly-used real-world data, revealing the current data scarcity for 3D point cloud registration.

\section{Conclusion}
We present \ourname{}, a novel 3D point cloud registration framework that boosts models with generative data for improved performance. In \ourname{}, a real depth map is re-projected with a random transformation and inpainted with a pre-trained diffusion model to generate a partially overlapped point cloud pair. Range-null space decomposition and our well-trained \ourdiffusionname{} ensure 3D consistency in overlap regions and realism of new content. This is the first approach that automatically generates realistic 3D point cloud registration training data using diffusion models. Besides, our \ourdepthcorretionname{} addresses the point penetration, avoiding artifacts and ensuring high-quality data generation. Experimental results confirm the state-of-the-art performance and robustness of \ourname{} on both indoor and outdoor benchmarks.

\clearpage

\section*{Acknowledgments}
This work was supported by the National Natural Science Foundation of China (NSFC) under grants Nos.62071097 and 62372091, the Sichuan Science and Technology Program of China under grants Nos.2023NSFSC0458 and 2023NSFSC0462, the Research Grants Council of the Hong Kong Special Administrative Region, China (Project Reference Number T45-401/22-N), and the Project of Hetao Shenzhen-Hong Kong Science and Technology Innovation Cooperation Zone (HZQB-KCZYB-2020083).

\bibliographystyle{splncs04}
\bibliography{main}

\begin{thebibliography}{10}
\providecommand{\url}[1]{\texttt{#1}}
\providecommand{\urlprefix}{URL }
\providecommand{\doi}[1]{https://doi.org/#1}

\bibitem{Ao_2023_CVPR}
Ao, S., Hu, Q., Wang, H., Xu, K., Guo, Y.: {BUFFER}: Balancing accuracy, efficiency, and generalizability in point cloud registration. In: CVPR. pp. 1255--1264 (June 2023)

\bibitem{ao2021spinnet}
Ao, S., Hu, Q., Yang, B., Markham, A., Guo, Y.: {SpinNet}: Learning a general surface descriptor for {3D} point cloud registration. In: CVPR. pp. 11753--11762 (2021)

\bibitem{aoki2019pointnetlk}
Aoki, Y., Goforth, H., Srivatsan, R.A., Lucey, S.: {PointNetLK}: Robust \& efficient point cloud registration using {PointNet}. In: CVPR. pp. 7163--7172 (2019)

\bibitem{bai2021pointdsc}
Bai, X., Luo, Z., Zhou, L., Chen, H., Li, L., Hu, Z., Fu, H., Tai, C.L.: {PointDSC}: Robust point cloud registration using deep spatial consistency. In: CVPR. pp. 15859--15869 (2021)

\bibitem{bai2020d3feat}
Bai, X., Luo, Z., Zhou, L., Fu, H., Quan, L., Tai, C.L.: {D3Feat}: Joint learning of dense detection and description of {3D} local features. In: CVPR. pp. 6359--6367 (2020)

\bibitem{besl1992method}
Besl, P.J., McKay, N.D.: A method for registration of {3D} shapes. IEEE TPAMI  \textbf{14}(2),  239--256 (1992)

\bibitem{Chen_2023_ICCV}
Chen, G., Wang, M., Yuan, L., Yang, Y., Yue, Y.: Rethinking point cloud registration as masking and reconstruction. In: ICCV. pp. 17717--17727 (October 2023)

\bibitem{chen2023sira}
Chen, S., Xu, H., Li, R., Liu, G., Fu, C.W., Liu, S.: {SIRA-PCR}: Sim-to-real adaptation for {3D} point cloud registration. In: ICCV. pp. 14394--14405 (October 2023)

\bibitem{choy2020deep}
Choy, C., Dong, W., Koltun, V.: Deep global registration. In: CVPR. pp. 2514--2523 (2020)

\bibitem{choy2019fully}
Choy, C., Park, J., Koltun, V.: Fully convolutional geometric features. In: ICCV. pp. 8958--8966 (2019)

\bibitem{Dang_2023_ICCV}
Dang, Z., Salzmann, M.: {AutoSynth}: Learning to generate {3D} training data for object point cloud registration. In: ICCV. pp. 9009--9019 (October 2023)

\bibitem{dhariwal2021diffusion}
Dhariwal, P., Nichol, A.: Diffusion models beat {GAN}s on image synthesis. NeurIPS  \textbf{34},  8780--8794 (2021)

\bibitem{fischler1981random}
Fischler, M.A., Bolles, R.C.: Random sample consensus: A paradigm for model fitting with applications to image analysis and automated cartography. Communications of the ACM  \textbf{24}(6),  381--395 (1981)

\bibitem{fu2021robust}
Fu, K., Liu, S., Luo, X., Wang, M.: Robust point cloud registration framework based on deep graph matching. In: CVPR. pp. 8893--8902 (2021)

\bibitem{gojcic2019perfect}
Gojcic, Z., Zhou, C., Wegner, J.D., Wieser, A.: The perfect match: {3D} point cloud matching with smoothed densities. In: CVPR. pp. 5545--5554 (2019)

\bibitem{Hatem_2023_ICCV}
Hatem, A., Qian, Y., Wang, Y.: {Point-TTA}: Test-time adaptation for point cloud registration using multitask meta-auxiliary learning. In: ICCV. pp. 16494--16504 (October 2023)

\bibitem{ho2020denoising}
Ho, J., Jain, A., Abbeel, P.: Denoising diffusion probabilistic models. NeurIPS  \textbf{33},  6840--6851 (2020)

\bibitem{ho2022classifier}
Ho, J., Salimans, T.: Classifier-free diffusion guidance. arXiv preprint arXiv:2207.12598  (2022)

\bibitem{horache20213d}
Horache, S., Deschaud, J.E., Goulette, F.: {3D} point cloud registration with multi-scale architecture and unsupervised transfer learning. In: 3DV. pp. 1351--1361. IEEE (2021)

\bibitem{abs-2106-10859}
Hsu, C., Sun, C., Chen, H.: Moving in a 360 world: Synthesizing panoramic parallaxes from a single panorama. CoRR  \textbf{abs/2106.10859} (2021), \url{https://arxiv.org/abs/2106.10859}

\bibitem{huang2020predator}
Huang, S., Gojcic, Z., Usvyatsov, M., Wieser, A., Schindler, K.: {PREDATOR}: Registration of {3D} point clouds with low overlap. arXiv:2011.13005  (2020)

\bibitem{huang2020feature}
Huang, X., Mei, G., Zhang, J.: Feature-metric registration: A fast semi-supervised approach for robust point cloud registration without correspondences. In: CVPR. pp. 11366--11374 (2020)

\bibitem{jabri2022scalable}
Jabri, A., Fleet, D., Chen, T.: Scalable adaptive computation for iterative generation. arXiv preprint arXiv:2212.11972  (2022)

\bibitem{Jiang_2023_CVPR}
Jiang, H., Dang, Z., Wei, Z., Xie, J., Yang, J., Salzmann, M.: Robust outlier rejection for {3D} registration with variational bayes. In: CVPR. pp. 1148--1157 (June 2023)

\bibitem{Karras2022edm}
Karras, T., Aittala, M., Aila, T., Laine, S.: Elucidating the design space of diffusion-based generative models. In: NeurIPS (2022)

\bibitem{lei2023rgbd2}
Lei, J., Tang, J., Jia, K.: {RGBD2}: Generative scene synthesis via incremental view inpainting using {RGBD} diffusion models. In: CVPR. pp. 8422--8434 (2023)

\bibitem{li2024dmhomo}
Li, H., Jiang, H., Luo, A., Tan, P., Fan, H., Zeng, B., Liu, S.: {DMHomo}: Learning homography with diffusion models. ACM TOG  \textbf{43}(3),  1--16 (2024)

\bibitem{idam}
Li, J., Zhang, C., Xu, Z., Zhou, H., Zhang, C.: Iterative distance-aware similarity matrix convolution with mutual-supervised point elimination for efficient point cloud registration. In: ECCV. pp. 378--394 (2020)

\bibitem{lepard2021}
Li, Y., Harada, T.: Lepard: Learning partial point cloud matching in rigid and deformable scenes. CVPR  (2022)

\bibitem{liu2023regformer}
Liu, J., Wang, G., Liu, Z., Jiang, C., Pollefeys, M., Wang, H.: {RegFormer}: An efficient projection-aware transformer network for large-scale point cloud registration. In: ICCV. pp. 8451--8460 (2023)

\bibitem{Liu_2023_ICCV}
Liu, Q., Zhu, H., Zhou, Y., Li, H., Chang, S., Guo, M.: Density-invariant features for distant point cloud registration. In: ICCV. pp. 18215--18225 (October 2023)

\bibitem{liu2023zero}
Liu, R., Wu, R., Van~Hoorick, B., Tokmakov, P., Zakharov, S., Vondrick, C.: Zero-1-to-3: Zero-shot one image to {3D} object. In: CVPR. pp. 9298--9309 (2023)

\bibitem{liu2023more}
Liu, X., Park, D.H., Azadi, S., Zhang, G., Chopikyan, A., Hu, Y., Shi, H., Rohrbach, A., Darrell, T.: More control for free! image synthesis with semantic diffusion guidance. In: WACV. pp. 289--299 (2023)

\bibitem{luo2023image}
Luo, Z., Gustafsson, F.K., Zhao, Z., Sj{\"o}lund, J., Sch{\"o}n, T.B.: Image restoration with mean-reverting stochastic differential equations. arXiv preprint arXiv:2301.11699  (2023)

\bibitem{Mei_2023_CVPR}
Mei, G., Tang, H., Huang, X., Wang, W., Liu, J., Zhang, J., Van~Gool, L., Wu, Q.: Unsupervised deep probabilistic approach for partial point cloud registration. In: CVPR. pp. 13611--13620 (June 2023)

\bibitem{pais20203dregnet}
Pais, G.D., Ramalingam, S., Govindu, V.M., Nascimento, J.C., Chellappa, R., Miraldo, P.: {3DRegNet}: A deep neural network for {3D} point registration. In: CVPR. pp. 7193--7203 (2020)

\bibitem{poiesi2022learning}
Poiesi, F., Boscaini, D.: Learning general and distinctive {3D} local deep descriptors for point cloud registration. IEEE TPAMI  (2022)

\bibitem{pomerleau2012challenging}
Pomerleau, F., Liu, M., Colas, F., Siegwart, R.: Challenging data sets for point cloud registration algorithms. IJRR  \textbf{31}(14),  1705--1711 (2012)

\bibitem{qin2022geometric}
Qin, Z., Yu, H., Wang, C., Guo, Y., Peng, Y., Xu, K.: Geometric transformer for fast and robust point cloud registration. In: CVPR. pp. 11143--11152 (June 2022)

\bibitem{rombach2022high}
Rombach, R., Blattmann, A., Lorenz, D., Esser, P., Ommer, B.: High-resolution image synthesis with latent diffusion models. In: CVPR. pp. 10684--10695 (2022)

\bibitem{rusinkiewicz2019symmetric}
Rusinkiewicz, S.: A symmetric objective function for {LiDAR}. ACM TOG  \textbf{38}(4), ~1--7 (2019)

\bibitem{rusinkiewicz-normal-sampling}
Rusinkiewicz, S., Levoy, M.: Efficient variants of the {ICP} algorithm. In: International Conference on 3-D Digital Imaging and Modeling (3DIM). pp. 145--152 (2001)

\bibitem{FPFH}
Rusu, R.B., Blodow, N., Beetz, M.: Fast point feature histograms ({FPFH}) for {3D} registration. In: ICRA. pp. 3212--3217 (2009)

\bibitem{saharia2022image}
Saharia, C., Ho, J., Chan, W., Salimans, T., Fleet, D.J., Norouzi, M.: Image super-resolution via iterative refinement. IEEE TPAMI  (2022)

\bibitem{segal2009generalized}
Segal, A., Haehnel, D., Thrun, S.: Generalized-{LiDAR}. In: Robotics: Science and Systems. vol.~2, p.~435 (2009)

\bibitem{sohl2015deep}
Sohl-Dickstein, J., Weiss, E., Maheswaranathan, N., Ganguli, S.: Deep unsupervised learning using nonequilibrium thermodynamics. In: ICML. pp. 2256--2265. PMLR (2015)

\bibitem{song2020denoising}
Song, J., Meng, C., Ermon, S.: Denoising diffusion implicit models. arXiv preprint arXiv:2010.02502  (2020)

\bibitem{song2021solving}
Song, Y., Shen, L., Xing, L., Ermon, S.: Solving inverse problems in medical imaging with score-based generative models. arXiv preprint arXiv:2111.08005  (2021)

\bibitem{tevet2022human}
Tevet, G., Raab, S., Gordon, B., Shafir, Y., Cohen-Or, D., Bermano, A.H.: Human motion diffusion model. arXiv preprint arXiv:2209.14916  (2022)

\bibitem{wang2022you}
Wang, H., Liu, Y., Dong, Z., Wang, W.: You only hypothesize once: Point cloud registration with rotation-equivariant descriptors. In: IEEE TMM. pp. 1630--1641 (2022)

\bibitem{wang2023roreg}
Wang, H., Liu, Y., Hu, Q., Wang, B., Chen, J., Dong, Z., Guo, Y., Wang, W., Yang, B.: {RoReg}: Pairwise point cloud registration with oriented descriptors and local rotations. IEEE TPAMI  (2023)

\bibitem{wang2022zero}
Wang, Y., Yu, J., Zhang, J.: Zero-shot image restoration using denoising diffusion null-space model. arXiv preprint arXiv:2212.00490  (2022)

\bibitem{wang2019deep}
Wang, Y., Solomon, J.M.: Deep closest point: Learning representations for point cloud registration. In: ICCV. pp. 3523--3532 (2019)

\bibitem{wang2019prnet}
Wang, Y., Solomon, J.M.: {PRNet}: Self-supervised learning for partial-to-partial registration. In: NeurIPS. pp. 8814--8826 (2019)

\bibitem{xu2024handbooster}
Xu, H., Li, H., Wang, Y., Liu, S., Fu, C.W.: {HandBooster}: Boosting {3D} hand-mesh reconstruction by conditional synthesis and sampling of hand-object interactions. In: CVPR. pp. 10159--10169 (2024)

\bibitem{xu2021omnet}
Xu, H., Liu, S., Wang, G., Liu, G., Zeng, B.: {OMNet}: Learning overlapping mask for partial-to-partial point cloud registration. In: ICCV (2021)

\bibitem{xu2022finet}
Xu, H., Ye, N., Liu, G., Zeng, B., Liu, S.: {FINet}: Dual branches feature interaction for partial-to-partial point cloud registration. In: AAAI. vol.~36, pp. 2848--2856 (2022)

\bibitem{yangone}
Yang, F., Guo, L., Chen, Z., Tao, W.: One-inlier is first: Towards efficient position encoding for point cloud registration. NeurIPS  \textbf{35},  6982--6995 (2022)

\bibitem{yang2013go}
Yang, J., Li, H., Jia, Y.: {Go-ICP}: Solving {3D} registration efficiently and globally optimally. In: CVPR. pp. 1457--1464 (2013)

\bibitem{yang2024single}
Yang, Z., Li, H., Hong, M., Zeng, B., Liu, S.: Single image rolling shutter removal with diffusion models. arXiv preprint arXiv:2407.02906  (2024)

\bibitem{yew2020-RPMNet}
Yew, Z.J., Lee, G.H.: {RPM-Net}: Robust point matching using learned features. In: CVPR. pp. 11824--11833 (2020)

\bibitem{yew2022regtr}
Yew, Z.J., Lee, G.H.: {REGTR}: End-to-end point cloud correspondences with transformers. In: CVPR. pp. 6677--6686 (2022)

\bibitem{yu2021cofinet}
Yu, H., Li, F., Saleh, M., Busam, B., Ilic, S.: {CoFiNet}: Reliable coarse-to-fine correspondences for robust point cloud registration. NeurIPS  \textbf{34},  23872--23884 (2021)

\bibitem{yu2023peal}
Yu, J., Ren, L., Zhang, Y., Zhou, W., Lin, L., Dai, G.: {PEAL}: Prior-embedded explicit attention learning for low-overlap point cloud registration. In: CVPR. pp. 17702--17711 (2023)

\bibitem{zeng20173dmatch}
Zeng, A., Song, S., Nie{\ss}ner, M., Fisher, M., Xiao, J., Funkhouser, T.: {3DMatch}: Learning local geometric descriptors from {RGB-D} reconstructions. In: CVPR. pp. 1802--1811 (2017)

\bibitem{Zhang_2023_CVPR}
Zhang, X., Yang, J., Zhang, S., Zhang, Y.: {3D} registration with maximal cliques. In: CVPR. pp. 17745--17754 (June 2023)

\bibitem{zhou2024recdiffusion}
Zhou, T., Li, H., Wang, Z., Luo, A., Zhang, C.L., Li, J., Zeng, B., Liu, S.: {RecDiffusion}: Rectangling for image stitching with diffusion models. In: CVPR. pp. 2692--2701 (2024)

\end{thebibliography}
\end{document}